\documentclass[10pt,twocolumn,letterpaper]{article}

\usepackage{iccv}              

\definecolor{iccvblue}{rgb}{0.21,0.49,0.74}
\usepackage[pagebackref,breaklinks,colorlinks,allcolors=iccvblue]{hyperref}
\usepackage{kotex}
\usepackage{comment}
\newcommand{\sy}[1]{\textcolor{black}{#1}}
\newcommand{\yr}[1]{\textcolor{black}{#1}}

\def\paperID{00006} 
\def\confName{ICCV Workshop(exCV Non-proceeding)}
\def\confYear{2025}

\title{Causal Interpretation of Sparse Autoencoder Features in Vision}

\author{
Sangyu Han \quad Yearim Kim \quad Nojun Kwak\\
Seoul National University, Seoul, Korea\\
{\tt\small \{acoexist96, yerim1656, nojunk\}@snu.ac.kr}
}

\begin{document}
\maketitle
\begin{abstract}
Understanding what sparse auto-encoder (SAE) features in vision transformers truly represent is usually done by inspecting the patches where a feature’s activation is highest. However, self-attention mixes information across the entire image, so an activated patch often co-occurs with—but does not cause—the feature’s firing. We propose \textbf{Causal Feature Explanation (CaFE)}, which levarages Effective Receptive Field (ERF). 
We consider each activation of an SAE feature to be a target and apply input-attribution methods to identify the image patches that causally drive that activation.
Across CLIP-ViT features, ERF maps frequently diverge from naive activation maps, revealing hidden context dependencies (e.g., a “roaring face” feature that requires \yr{the co-occurrence of} eyes and nose, \yr{rather than merely an open mounth.}). Patch insertion tests confirm that our CaFE more effectively recovers or suppresses feature activations than activation-ranked patches. Our results show that CaFE yields more faithful and semantically precise explanations of vision-SAE features, highlighting the risk of misinterpretation when relying solely on activation location.

\end{abstract}    
\section{Introduction}
\label{sec:intro}


\begin{figure}[ht]
    \centering
    \includegraphics[width=.8\columnwidth]{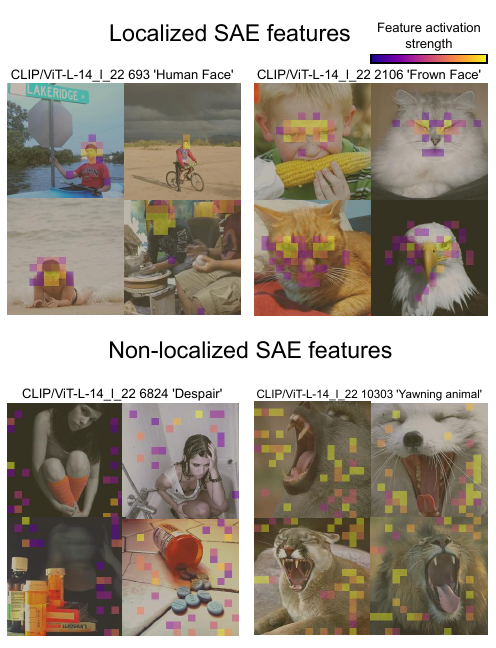}
    \caption{While most SAE feature activations are localized aligned with its meaning~(\textbf{Top}), We found some of SAE feature activations scattered across the image, halting the explanation of SAE features~(\textbf{Bottom}).} 
    \label{fig:intro}
\end{figure}
\begin{figure*}[ht]
    \centering
    \includegraphics[width=0.9\textwidth]{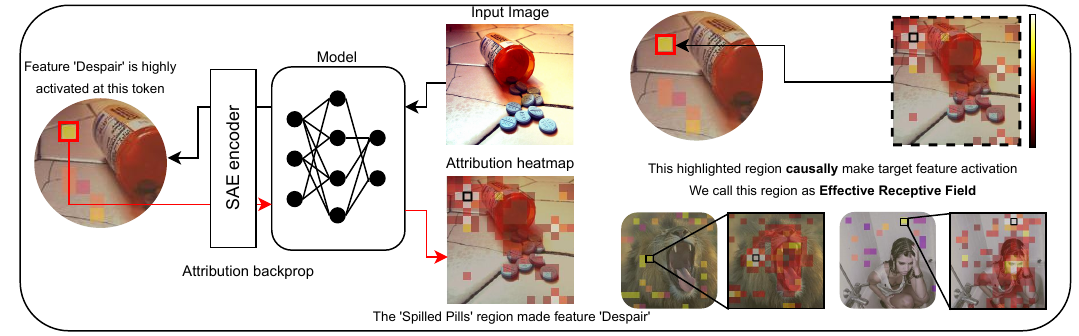}
    \caption{Method overview. The feature 'Despair' highly activated at the patch at background. Using attribution method, we can find which part of image causally contribute to the feature. We call this effective region of interest as \textbf{Effective Receptive Field.}   }
    \label{fig:overview}
\end{figure*}

Interpretable machine learning seeks to map deep model representations to human-understandable concepts. 
In vision, sparse autoencoders (SAEs) have emerged as a compelling approach to distill concise basis features from high-dimensional transformer latents by imposing sparsity constraints~\cite{huben2024sparse}.
These SAE features ideally capture distinct visual patterns and have been widely used to reveal semantic structure in complex models.

To assign semantic labels to these SAE features, prior works have proposed several methodologies. 
One line of work retrieves the top-activated samples and derives feature labels from those images~\cite{zaigrajew2025interpreting, pach2025saevlm}.
Another approach further pinpoints high-activation tokens within those samples for fine-grained annotation \cite{stevens2025scientificallyrigorous}.
More recently, PatchSAE combines both sample- and patch-level analyses, though it still primarily relies on top-activated images for interpretation \cite{lim2025sparse}.
All these approaches presuppose that features are spatially localized and semantically coherent.

However, as shown in Fig.~\ref{fig:intro}, we find that there exists some sparse autoencoder (SAE) features 
are \emph{non-localized}: their highest-activation patches scatter across the image.
For these non-localized features, neither reviewing top-activated images nor marking their activated patches yields a coherent interpretation, since the visible activation peaks represent correlational, not necessarily causal, evidence.

To interpret non-localized SAE features faithfully, we propose examining each activated token’s \emph{effective receptive field} (ERF)~\cite{han2024respect}: the set of input patches that causally drive the token’s activation.
Only by tracing attribution beyond the spatial location of maximum activation can one uncover the true image evidence behind a feature.

In this paper, we introduce \textbf{Causal Feature Explanation (CaFE)}, which integrates: (1) SAE feature extraction on transformer latents, and (2) patch-level input attribution (e.g., Integrated Gradients or AttnLRP).
For each feature, we compute attribution scores over all patches and identify those with highest causal contribution, forming the feature’s ERF.
By revealing the causal drivers of feature activations, CaFE enables more trustworthy and accurate interpretations of vision models, avoiding misleading correlations and deepening our understanding of complex architectures.

\section{Related works}
\label{sec:related}

Individual units within convolutional neural networks (CNN) and transformer architectures are often \emph{polysemantic}, entangling several concepts and thereby hampering interpretation.
Sparse autoencoders (SAEs) address this limitation by learning an overcomplete, $l_1$-regularized basis in which each latent dimension activates for a single concept.
Originally applied to language models, SAEs have seen adapted to vision transformers, where they capture object parts and textures from patch embeddings~\cite{lim2025sparse}.
Most existing explanation methods simply visualize top-activating images or tokens~\cite{zaigrajew2025interpreting, stevens2025scientificallyrigorous}; these works are effective only when features are spatially localized and may fail when confronted with the non-localized features we observe.

\section{Method}
\begin{figure*}[ht]
    \centering
    \includegraphics[width=\textwidth]{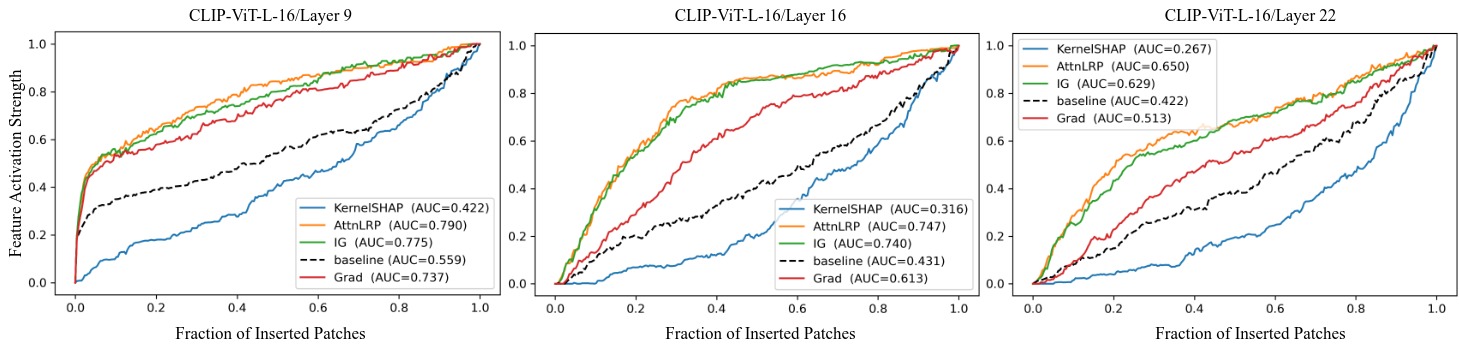}
        \caption{Causality validation. We compare causality of our CaFE method with baseline (naive activation-based patch ranking). We also compare various attribution methods including KernelSHAP, Attention-LRP, Integrated Gradients, and Gradients. \sy{It shows that our CaFE method with AttnLRP surpassed other attribution methods.
        }} 
    \label{fig:insertion}
\end{figure*}

\subsection{Preliminaries: SAE Features}
Given a hidden representation $\mathbf{h}\in\mathbb{R}^{n}$ from a backbone (e.g.\ patch embeddings of a ViT), we learn a SAE
\[
\mathbf{z}=\textrm{ReLU}(W_e(\mathbf{h}-\mathbf{b}_d)), \qquad
\hat{\mathbf{h}}=W_d\mathbf{z} - \mathbf{b}_h,
\]
with the number of features $m \gg n$, where $W_e \in \mathbb{R}^{m \times d}$ is the SAE encoder weight matrix, $W_d \in \mathbb{R}^{d \times m}$ is the decoder weight matrix, and $\mathbf{b}_d, \mathbf{b}_h$ are learned bias vectors. Training minimizes
\[
\mathcal{L}
=\bigl\lVert \mathbf{h}-\hat{\mathbf{h}}\bigr\rVert_2^{2}
+\lambda\lVert \mathbf{z}\rVert_{1},
\]
so that each latent dimension $z_k$ activates only for patches exhibiting a specific visual concept.  The $k$-th row of $W_e$ therefore serves as an interpretable feature detector, often corresponding to an object part, texture, or scene element.

\subsection{Causal Feature Explanation~(CaFE)}
\label{sec:causal_feature}
\paragraph{From activations to causes.} 
SAE features are usually inspected by ranking the patches whose activations $z_k(I) \in \mathbb{R}$ are highest.
While this works well for \emph{localized} features, Fig ~\ref{fig:intro}
 shows that \emph{non-localized} features appear in disparate regions, offering only correlational hints. 
 This limitation motivates us to ask not merely \emph{where} $z_k$ fires, but \emph{which image evidence truly drives the activation}.

\paragraph{Effective Receptive Field (ERF).}  
Let $A(p\mid z_k, I)$ denote the attribution of \sy{input} patch $p$ to the scalar output $z_k(I)$.  
The \emph{effective receptive field} of feature $k$ on $I$ is the score map

\begin{equation}
    \text{ERF}_k(I) = \bigl\{\bigl(p,A(p\mid~z_k,I)\bigr) :p\in I\bigr\},
\end{equation}
whose intensity measures how much $p$ caused to form $z_k$.
High‑score regions are the real evidence behind the activation, independent of where the network reports the maximum.  
Fig.~\ref{fig:overview} illustrates that although the token in the background floor attains the highest activation for the “\textit{Despair}” feature, the ERF pinpoints the spilled‑pill region as the causal driver.

As shown in Fig.~\ref{fig:overview}, the attribution $A$ is obtanined by backpropagating relevance scores from the target SAE neuron through the SAE encoder and subsequently through the vision transformer.
Inside transformer layers, we employ \textbf{Attention‑LRP (AttnLRP)}~\cite{achtibat2024attnlrp}, an adaptation of Layer‑wise Relevance Propagation that distributes relevance \sy{properly considering} attention edges.
For comparison, we also report \emph{Integrated Gradients} (IG)~\cite{sundararajan2017axiomatic}, KernelSHAP~\cite{lundberg2017unified}, and Gradients as baselines. 
These methods produce the same ERF interface, allowing plug‑and‑play replacements.

By replacing naive activation maps with ERF attribution, our \textbf{Causal Feature Explanation (CaFE)} pipeline delivers faithful explanations for both localized and non‑localized SAE features, laying the foundation for the analyses in the following sections.

\section{Experiments}


\begin{figure*}[ht]
    \centering
    \includegraphics[width=\textwidth]{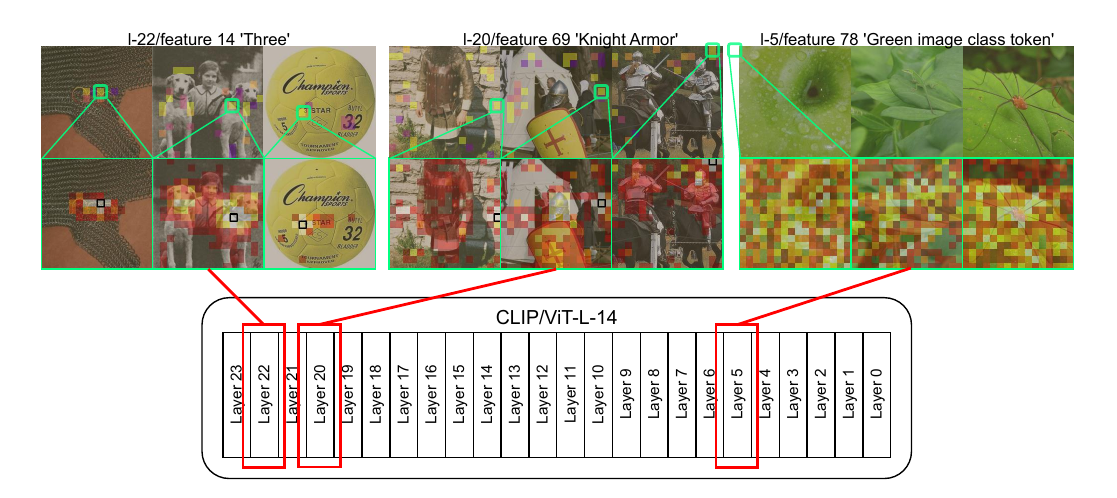}
    \caption{Qualitative examples of non-local SAE features and their ERFs at the points of highest activation across different layers. Even when the feature is spatially displaced from the region that encodes its meaning, its ERF still correctly pinpoints the area that triggered the activation. In lower layers~(right-most column), the non-local SAE feature appears only when its semantic meaning is linked to class tokens.
    }
    \label{fig:qualitative}
\end{figure*}

\begin{figure}[]
    \centering
	\includegraphics[width=.9\columnwidth]{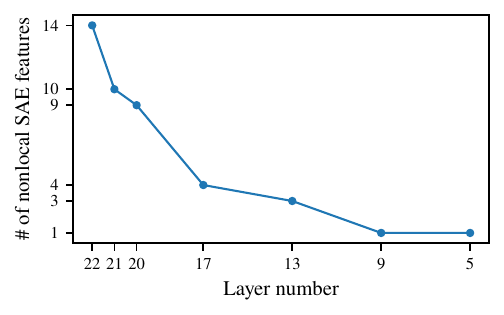}
    \caption{\yr{The number of non-local SAE features across layers.} The non-local SAE features become rarer as layer number decreases. 
    We manually inspected the number of non-local features out of the first 100 features. 
    }
    \label{fig:layerwise}
\end{figure}

We conduct a series of experiments to (i)~assess the fidelity with which \yr{CaFE} identifies the \emph{causal} image evidence behind SAE activations and (ii)~characterize when and where non-local SAE features arise in a vision transformer.

\paragraph{Experimental Setup.}
All experiments are conducted with the CLIP--ViT-L/14 encoder released by~\cite{cherti2023reproducible}.  
For each transformer layer, we train a Matryoshka SAE~\cite{huben2024sparse} on \textbf{5\,$\times$\,10\textsuperscript{8}} image patches extracted from the ImageNet-1K training set.  
Following the prior works, the reconstruction and sparsity hyper-parameters are held fixed across layers.

\subsection{Quantitative Causality Evaluation}
\label{subsec:insertion}
To quantitatively validate that our Causal Feature Explanation (CaFE) framework accurately identifies the true causal regions, we perform insertion tests, an established evaluation protocol in the explainability evaluations~\cite{samek2016evaluating,achtibat2024attnlrp,han2024respect}. 
In an insertion test, we start with a blank image and gradually insert patches from the original image in order of their importance, measuring how quickly the feature activation is recovered.
We then apply these insertion tests to compare the efficacy of attribution-based importance maps with that of the naive activation-based patch ranking.
If \yr{our method with ERF} truly pinpoints the true causal patches, then inserting solely those patches into a blank canvas should yield a higher $z_k$ activation than inserting the patches selected purely on the basis of their activation magnitudes. 
We perform these tests over a set of images and features, computing metrics like the area under the insertion curve (AUC) as a summary of explanation efficacy.
\sy{We compared several CaFE methods with different attribution with baseline approach. } 
Fig.~\ref{fig:insertion} confirms that ERF-guided insertion highly outperformed the causal recovery rate of the activation baseline across the layers.
\sy{Between CaFE methods, AttnLRP surpassed other attribution methods - which is known as the most faithful attribution method for the transformer architecture.}
Importantly, even for traditional ``local" features ERF still gives a modest boost, indicating that attribution sharpens patch selection beyond raw activation.

\subsection{Qualitative Analysis of Non-local Features}
\label{subsec:qualitative}

ERF maps of non-local features frequently diverge from the naive activation map, uncovering hidden context dependencies.
\textbf{Bottom right} of fig.~\ref{fig:overview} shows the examples of non-local SAE features with their corresponding ERF.
The ``Despair'' feature fires only when \emph{spilled pills} co-occur with a \emph{frowning face}, even though its maximal activation patch is far from the pills.  
Such phenomena remain imperceptible when one relies solely on activation-based inspection. 

Moreover, for each layer, we manually reviewed the \textbf{first 100} SAE features and flagged those whose top-activation patches were spatially inconsistent with their corresponding ERF. 
The results depicted in Fig.~\ref{fig:layerwise} show a clear trend.
Firstly, non-local features are \emph{scarce} in early layers (\(<\)layer\,9).
All such cases are class-token features whose activations reside exclusively at the CLS position.
Secondly, their frequency rises sharply in higher layers, peaking at layer\,22 where \(\approx\)14\% of features are non-local.
These features encode highly abstract, often compositional concepts (e.g.\ ``knight in armour'', ``three'').

This distribution supports the intuition that self-attention progressively mixes global context, making later-layer activations increasingly difficult to interpret without ERF. 


\subsection{Discussion}
\label{subsec:discussion}


\paragraph{Limitations and open questions.}
Computing the ERF for each feature requires both forward and backward passes, which may be costly for very deep SAEs.  
Furthermore, manual annotation of non-local features is inherently subjective; devising scalable, automated criteria remains future work.  
Finally, while we focus on vision transformers, analogous patterns of context mixing arise in large-scale language models; extending our CaFE with ERF-based attribution to text modalities is also a compelling avenue for further exploration.

\section{Conclusion}
We presented the Causal Feature Explanation~(CaFE) framework for interpreting vision model features by leveraging effective receptive field attribution, and empircally demonstrated its superiority to the conventional practice of relying on top activations.
Our study reveals that many sparse autoencoder features in vision transformers cannot be adequately understood by looking only at where they activate; one must also consider why they activate.
By applying input attribution to each feature, we obtain causal explanations that often differ from correlational activation maps – shedding light on context dependencies and complex concepts encoded by the model.
The main contribution of our work is to show that causal, ERF-based explanations provide a more faithful and semantically precise interpretation of visual features than activation-based methods.
This is a crucial diagnostic insight for vision model interpretability: it prevents us from mislabeling features or overlooking the true factors influencing model representations.

\newpage
{
    \small
    \bibliographystyle{ieeenat_fullname}
    \bibliography{main}

\begin{thebibliography}{11}
\providecommand{\natexlab}[1]{#1}
\providecommand{\url}[1]{\texttt{#1}}
\expandafter\ifx\csname urlstyle\endcsname\relax
  \providecommand{\doi}[1]{doi: #1}\else
  \providecommand{\doi}{doi: \begingroup \urlstyle{rm}\Url}\fi

\bibitem[Achtibat et~al.(2024)Achtibat, Hatefi, Dreyer, Jain, Wiegand,
  Lapuschkin, and Samek]{achtibat2024attnlrp}
Reduan Achtibat, Sayed Mohammad~Vakilzadeh Hatefi, Maximilian Dreyer, Aakriti
  Jain, Thomas Wiegand, Sebastian Lapuschkin, and Wojciech Samek.
\newblock Attnlrp: attention-aware layer-wise relevance propagation for
  transformers.
\newblock \emph{arXiv preprint arXiv:2402.05602}, 2024.

\bibitem[Cherti et~al.(2023)Cherti, Beaumont, Wightman, Wortsman, Ilharco,
  Gordon, Schuhmann, Schmidt, and Jitsev]{cherti2023reproducible}
Mehdi Cherti, Romain Beaumont, Ross Wightman, Mitchell Wortsman, Gabriel
  Ilharco, Cade Gordon, Christoph Schuhmann, Ludwig Schmidt, and Jenia Jitsev.
\newblock Reproducible scaling laws for contrastive language-image learning.
\newblock In \emph{Proceedings of the IEEE/CVF Conference on Computer Vision
  and Pattern Recognition}, pages 2818--2829, 2023.

\bibitem[Han et~al.(2024)Han, Kim, and Kwak]{han2024respect}
Sangyu Han, Yearim Kim, and Nojun Kwak.
\newblock Respect the model: Fine-grained and robust explanation with sharing
  ratio decomposition.
\newblock \emph{arXiv preprint arXiv:2402.03348}, 2024.

\bibitem[Huben et~al.(2024)Huben, Cunningham, Smith, Ewart, and
  Sharkey]{huben2024sparse}
Robert Huben, Hoagy Cunningham, Logan~Riggs Smith, Aidan Ewart, and Lee
  Sharkey.
\newblock Sparse autoencoders find highly interpretable features in language
  models.
\newblock In \emph{The Twelfth International Conference on Learning
  Representations}, 2024.

\bibitem[Lim et~al.(2025)Lim, Choi, Choo, and Schneider]{lim2025sparse}
Hyesu Lim, Jinho Choi, Jaegul Choo, and Steffen Schneider.
\newblock Sparse autoencoders reveal selective remapping of visual concepts
  during adaptation.
\newblock In \emph{The Thirteenth International Conference on Learning
  Representations}, 2025.

\bibitem[Lundberg and Lee(2017)]{lundberg2017unified}
Scott~M Lundberg and Su-In Lee.
\newblock A unified approach to interpreting model predictions.
\newblock \emph{Advances in neural information processing systems}, 30, 2017.

\bibitem[Pach et~al.(2025)Pach, Karthik, Bouniot, Belongie, and
  Akata]{pach2025saevlm}
Mateusz Pach, Shyamgopal Karthik, Quentin Bouniot, Serge Belongie, and Zeynep
  Akata.
\newblock Sparse autoencoders learn monosemantic features in vision-language
  models, 2025.

\bibitem[Samek et~al.(2016)Samek, Binder, Montavon, Lapuschkin, and
  M{\"u}ller]{samek2016evaluating}
Wojciech Samek, Alexander Binder, Gr{\'e}goire Montavon, Sebastian Lapuschkin,
  and Klaus-Robert M{\"u}ller.
\newblock Evaluating the visualization of what a deep neural network has
  learned.
\newblock \emph{IEEE transactions on neural networks and learning systems},
  28\penalty0 (11):\penalty0 2660--2673, 2016.

\bibitem[Stevens et~al.(2025)Stevens, Chao, Berger-Wolf, and
  Su]{stevens2025scientificallyrigorous}
Samuel Stevens, Wei-Lun Chao, Tanya Berger-Wolf, and Yu Su.
\newblock Sparse autoencoders for scientifically rigorous interpretation of
  vision models, 2025.

\bibitem[Sundararajan et~al.(2017)Sundararajan, Taly, and
  Yan]{sundararajan2017axiomatic}
Mukund Sundararajan, Ankur Taly, and Qiqi Yan.
\newblock Axiomatic attribution for deep networks.
\newblock In \emph{International conference on machine learning}, pages
  3319--3328. PMLR, 2017.

\bibitem[Zaigrajew et~al.(2025)Zaigrajew, Baniecki, and
  Biecek]{zaigrajew2025interpreting}
Vladimir Zaigrajew, Hubert Baniecki, and Przemyslaw Biecek.
\newblock Interpreting {CLIP} with hierarchical sparse autoencoders.
\newblock In \emph{Forty-second International Conference on Machine Learning},
  2025.

\end{thebibliography}
}

\end{document}